\newif\ifdraft\draftfalse

\newif\ifappendix\appendixtrue

\documentclass[sigconf]{acmart}

\usepackage{algorithmic}
\usepackage{algorithm}
\usepackage{mathtools}
\usepackage{bbm}
\usepackage{graphicx}
\usepackage{multirow}
\usepackage{hhline}
\usepackage{tikz}

\newcommand*\circledNorm[1]{\tikz[baseline=(char.base)]{
    \node[shape=circle,draw,inner sep=1pt] (char) {#1};}}

\AtBeginDocument{%
  \providecommand\BibTeX{{%
      \normalfont B\kern-0.5em{\scshape i\kern-0.25em b}\kern-0.8em\TeX}}}

\definecolor{purple}{RGB}{112, 48, 160}

\copyrightyear{2022} 
\acmYear{2022} 
\setcopyright{acmlicensed}\acmConference[WWW '22 Companion]{Companion Proceedings of the Web Conference 2022}{April 25--29, 2022}{Virtual Event, Lyon, France}
\acmBooktitle{Companion Proceedings of the Web Conference 2022 (WWW '22 Companion), April 25--29, 2022, Virtual Event, Lyon, France}
\acmPrice{15.00}
\acmDOI{10.1145/3487553.3524923}
\acmISBN{978-1-4503-9130-6/22/04}

\begin{document}

\title{Rows from Many Sources: Enriching row completions from Wikidata with a pre-trained Language Model}

\author{Carina Negreanu}
\authornote{Both authors contributed equally to this research.}
\email{cnegreanu@microsoft.com}
\author{Alperen Karaoglu}
\authornotemark[1]
\affiliation{%
  \institution{Microsoft Research}
  \city{Cambridge}
  \country{UK}
}

\author{Jack Williams}
\affiliation{%
  \institution{Microsoft Research}
  \city{Cambridge}
  \country{UK}
\email{jack.williams@microsoft.com}}

\author{Shuang Chen}
\affiliation{%
  \institution{Harbin Institute of Technology}
  \city{Harbin}
  \country{China}}

\author{Daniel Fabian}
\affiliation{%
 \institution{Microsoft Research}
 \city{Cambridge}
 \country{UK}
 \email{daniel.fabian@microsoft.com}}

\author{Andrew Gordon}
\affiliation{%
  \institution{Microsoft Research}
  \city{Cambridge}
  \country{UK}
  \email{adg@microsoft.com}}

\author{Chin-Yew Lin}
\affiliation{%
  \institution{Microsoft Research}
  \city{Beijing}
  \country{China}
\email{cyl@microsoft.com}}

\renewcommand{\shortauthors}{Carina Negreanu, Alperen Karaoglu, et al.}

\begin{abstract}
  Row completion is the task of augmenting a given table of text and numbers with additional, relevant rows. The task divides into two steps: subject suggestion, the task of populating the main column; and gap filling, the task of populating the remaining columns. We present state-of-the-art results for subject suggestion and gap filling measured on a standard benchmark (WikiTables).

  Our idea is to solve this task by harmoniously combining knowledge base table interpretation and free text generation. We interpret the table using the knowledge base to suggest new rows and generate metadata like headers through property linking. To improve candidate diversity, we synthesize additional rows using free text generation via GPT-3, and crucially, we exploit the metadata we interpret to produce better prompts for text generation. Finally, we verify that the additional synthesized content can be linked to the knowledge base or a trusted web source such as Wikipedia.
\end{abstract}



\maketitle

\section{Introduction}
\begin{figure}
  \includegraphics[width=0.75\columnwidth]{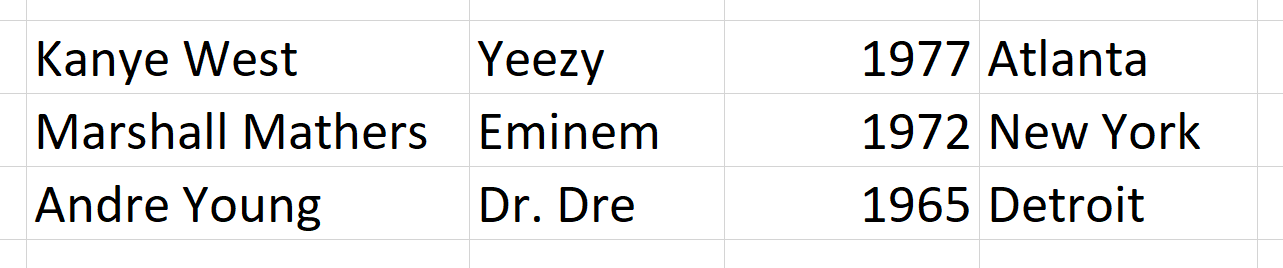}
  \\[2ex]
  \hrulefill
  \\[2ex]
  \includegraphics[width=0.75\columnwidth]{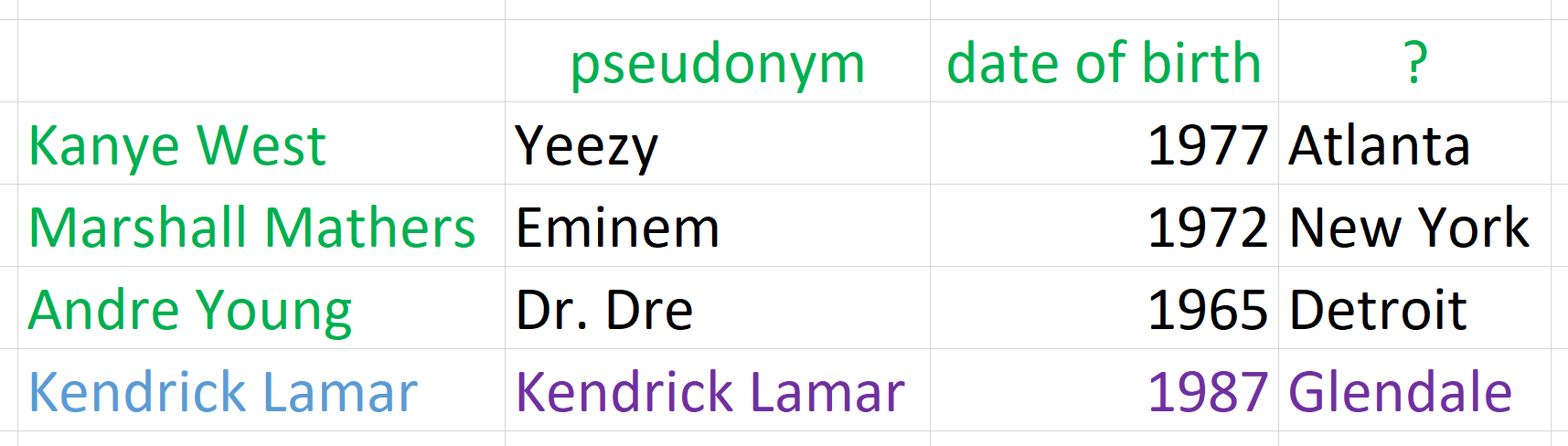}
  \Description[Example]{In a table about rappers the systems links the second column to pseudonyms, the third to date of birth and cannot link the forth. It suggests an extra row about Kendrick Lamar, with pseudonym Kendrick Lamar, date of birth 1987 and although not in Wikidata suggests Glendale as a potential value for the 4th column}
  \caption{Row Completion in a Nutshell}
  \label{fig:nutshell}
\end{figure}

\emph{Row completion} is the task of suggesting new complete rows for a given table.
Its purpose is to help users gain confidence in the completeness of their data.
Row completion is a table enhancement that is particularly important for spreadsheet intelligence, because it is a step towards unlocking a multitude of other intelligent features that require good quality data, such as data insights and chart recommendations. 

Row completion divides into three standard steps that we
illustrate using the example in Figure~\ref{fig:nutshell}.
To the best of our knowledge, the overall task of row completion is new, although each of its three steps has been studied independently.
The top table is the input table; the bottom table is the linked table with a
suggested row.
\begin{description}
\item[Step 1: Table Interpretation]
We link the data in the input table to a knowledge base (KB), such as DBpedia~\cite{auer2007dbpedia} or Wikidata~\cite{vrandevcic2014wikidata}.
Table interpretation itself decomposes into three sub-tasks as proposed in \cite{tabel}:
\textit{column type identification}, \textit{entity linking} and \textit{property linking}.
The figure shows in green the entities and properties from Wikidata (using the English labels as proxies for the unique numeric identifiers, and "?" denotes an unlinked property).

\item[Step 2: Subject Suggestion]
We predict additional primary entities described in the table, a task we refer to as \textit{subject suggestion}, since it determines the primary subject of the row and not the complete row.

Our figure shows in blue a single suggested subject: the entity ``Kendrick Lamar''.
In practice, subject suggestion generates an ordered set of entities.

\item[Step 3: Gap Filling]
We fill in the remainder of the row, a particular case of \textit{gap filling}.
Our figure shows in purple the properties of ``Kendrick Lamar'' filled by our algorithm.
\end{description}

The goal of this paper is to dramatically expand the scope of subject suggestion and gap filling by sourcing entities and properties from relatively uncurated web sources.
Our target is Wikidata, the largest, fastest growing, and most up to date open knowledge base.

To achieve this goal, we need to go beyond methods aimed at structured datasets like DBpedia or WikiTables, which inherit structure from their source Wikipedia, highly curated by humans.

\emph{Our key new idea is to rely on prompting an existing pre-trained language model during subject suggestion (Step 2) and gap filling (Step 3) to source entity and property candidates by free text generation.}

\subsubsection*{Challenges for Step 2: Subject Suggestion with Wikidata}

Subject suggestion first involves generating a set of potential candidates and then ranking them. EntiTables~\cite{entitables}, the current state of the art in candidate generation for subject suggestion, sources candidates from DBpedia and WikiTables. Although highly valuable, both sources are extracted from Wikipedia limiting the diversity.

To suggest from Wikidata we need to address two significant challenges. 
The first is that \emph{Wikidata has a weak type ontology.}

Existing approaches to subject suggestion, such as
Entitables~\cite{entitables}, rely on a well-defined type ontology to
construct search indexes to find candidates.
Unfortunately, Wikidata lacks such a well-defined type ontology, and
does not require users to provide complete information when entering a
new entity, leading to a significant number of missing properties~\cite{ontological-Wikidata}.
For example, the type of Cat and Lion in DBpedia is ``Mammal'', while in Wikidata,
a Cat is an instance of ``organisms known by a particular common
name'' (analogous to its type), and a Lion is an instance of ``taxon''.
We take a new approach where we start from multi-relational entity
embeddings (such as Pytorch BigGraph (PBG) embeddings \cite{BigG}) of
the seeds and create a candidate set from their nearest neighbours
that share a subset of properties linked from the table.  In doing so,
we can handle Wikidata's steep growth by avoiding expensive
computations over the whole entity space. Furthermore, PBG embeddings can be updated efficiently which makes them an ideal candidate for data that evolves rapidly over time.

The second challenge for subject suggestion from Wikidata is that \emph{directly using KB embeddings for subject suggestion can have low recall.} While Wikidata is vast, it remains incomplete which can lead to sparse 
regions in the embedding space as PBG trains on an input graph by ingesting its list of edges, each identified by its source and target entities and, possibly, a relation type. Thus, for some entities our initial approach might not have sufficient recall and we diversify our candidates by using free text generation. Generative language
models like GPT-3~\cite{GPT3} or Codex~\cite{codex} are trained on enormous datasets extracted from the
web, and capture a rich knowledge of co-occurrence
information. Augmenting knowledge base subject suggestion using a
language model is an enticing way to find new candidates, however
prompting GPT-3 using the input table alone delivers poor results.
Our approach is to exploit the table metadata we interpret in Step 1 to craft
structured prompts. For example, in Figure~\ref{fig:nutshell} we use
the linked properties to create the prompt ``Kanye West has pseudonym
Yeezy and has date of birth 1977''. Thus, we treat the large language model as a source of knowledge (that needs to be verified) which we query by synthesizing the tabular context. Our approach shows that good quality queries can be constructed for relational tables so we do not have to fine-tune the language model to understand the information in the table. 

Turning to candidate ranking for subject suggestion, prior art, such as Table2Vec~\cite{Li_Zhang} or more recently Tabbie~\cite{tabbie}, ranks via entity similarity (between seed entities and candidates) using knowledge base and table similarity statistics. They make use of tabular representations (static embeddings~\cite{Li_Zhang} or BERT-based representations~\cite{tabbie}) which require significant training and/or fine-tuning (e.g. ~\cite{tabbie} estimates total emissions at 300kg of $CO_2$ for this task). We propose building a classifier over features that are generated with metadata already available from the base models (e.g. GPT-3 and PBG) without further training or fine-tuning. As we have a large set of candidates relative to the number of suggestions we want to offer we classify using outlier detectors.

\subsubsection*{Challenges for Step 3: Gap Filling with Wikidata}
In Step 3, we want to fill in the remainder of the row once we have a subject suggestion. Recent state of the art for this task is achieved by TURL~\cite{deng2020turl}, a different table representation approach that learns contextualized representations on relational
tables by masked entity retrieval. It sources fills from other WikiTables.
In our work we fill gaps by using other sources such as Wikidata.
Unfortunately, KB sparsity directly affects gap filling as either the information we would like to retrieve doesn't exist (the entity's property value is missing) or there is not enough information to link the column to a property. To overcome this challenge we extend gap filling to include information from multiple sources such as news, Wikipedia articles or any other reliable web sources. 
Similarly to subject suggestion we use GPT-3 to generate potential candidates. The difference between the two approaches is that instead of trying to verify the candidates by linking back to Wikidata we attempt to ground the generations in a broader range of sources. Assigning provenance to a generation is insufficient to assess whether the generation should be a fill. We need to also use the context from the source to determine if the generation is consistent with the table. For example, if we had a table about athletes we could use GPT-3 to fill in the missing eye-colour information. If one of the generations is "brown" we need to verify that the colour refers to eye colour and not hair colour. 

The overall contribution of the paper is to establish a new state-of-the-art for row completion, and its key steps of subject suggestion and gap filling, by using language models in those two steps.

\section{Related Work}
 Various aspects of Row Completion have been previously studied and in this section we will provide an overview. 
 For candidate generation previous work has proposed sourcing candidates from a Knowledge Base such as DBpedia or Freebase (\cite{T2K}, \cite{egoset}, \cite{ESKG}, \cite{DBLP:conf/pakdd/ZhengSCLW17}, \cite{entitables}), or use free text (\cite{sarmento2007more}, \cite{shen2017setexpan}, \cite{DBLP:conf/ccpr/QiLZ12}) or use structured text such as web tables (\cite{wang2015concept}, \cite{deng2020turl}, \cite{entitables}) or web queries (\cite{gupta2009answering}, \cite{seisa}, \cite{DBLP:journals/tweb/XiaoLC20}). As several approaches we focus on multi source candidates and our work is closest to \cite{entitables} and \cite{Li_Zhang}.

When extracting candidates from KBs prior art favours DBpedia where comparing types and categories between all entities leads to good candidate generation. This approach works well for well curated KBs, but in the case of Wikidata we found that it is less effective considering its flexible ontology. To this point, recent work \cite{categ} proposes category generation for sets of entities.

  For ranking candidates, previous work uses various model classes, for example Machine Learning approaches (\cite{yan2020end}, \cite{deng2020turl}, \cite{LMSE}, \cite{DBLP:conf/acl/ZhangSSH20}, \cite{BSE}, \cite{tabbie}), probabilistic approaches (\cite{gupta2009answering}, \cite{wang2015concept}, \cite{DBLP:journals/tweb/XiaoLC20}, \cite{shen2017setexpan}, \cite{DBLP:conf/nips/ValeraG14}, \cite{entitables}, \cite{BS}), graph theory approaches (\cite{DBLP:conf/aaai/ZhangSH16}), and human-in-the-loop approaches (\cite{8332589}, \cite{10.1145/2791347.2791353}). Our pipeline uses Machine Learning components as we include free text generation and use the output of an unsupervised outlier detector as a ranking function. Unlike previous work we do not train or fine-tune the language model or the embeddings on tabular data. Due to the lack of a large corpus of varied data (the datasets are primarily synthetic or carefully curated) we are wary of generalization. Nonetheless recent deep learning tabular representation models (such as ~\cite{tabbie}, ~\cite{deng2020turl} or ~\cite{tabert}) show improvements by better encoding the tabular context. 

A related area of study is \emph{set expansion}, an instance of subject suggestion specialized to one column tables without headers, or lists.
Relevant work in this space that tackles ambiguity and semantic drift includes \cite{seisa}, \cite{egoset}, \cite{shen2017setexpan}, \cite{yu2019corpus}, \cite{ESKG}, \cite{sarmento2007more}, \cite{Li_Zhang} and \cite{DBLP:conf/ccpr/QiLZ12}. 

  For the gap filling task \cite{cac} aims to fill in gaps in tables with multi-source candidates (from DBpedia and tables) and their algorithms for tabular sources get improved upon by recent work \cite{deng2020turl}. Our work extends to unstructured text sources (like Wikipedia or news articles) for significant recall and precision gains. 

  Tangential emerging research areas that are relevant to our work are knowledge acquisition via pre-trained language models and prompt-engineering. Prior work (such as ~\cite{LMKB}, ~\cite{fact} or ~\cite{factual}) uses the knowledge within pre-trained language models for QA, fact checking or truthful generation. Significant efforts have focused on building better prompts and a representative collection can be found in a recent survey ~\cite{prompt_survey}. To our knowledge we are the first to focus on prompt engineering for tabular data.

\section{Tasks}

In this section, we formally define the tabular data interpretation and augmentation problem. We start by describing the tabular data and knowledge base (KB), then formally define the task settings.

Let $T$ range over relational tables of the form $\{t_{1,1},...,t_{m,n}\}$, where $t$ ranges over text values. A table is a matrix of text cells of $m$ rows and $n$ columns. Write $T_{[i,*]}$ for row $i$ in table $T$; write $T_{[*,j]}$ for column $j$ in table $T$. We do not assume availability of metadata beyond the table contents, such as column headers or types.

The knowledge base (KB) studied in our work follows RDF (Resource Description Framework) standard which consists of a terminological component (TBox) and an assertion component (ABox).
Let $\mathcal{E} = \{e_1,...,e_{|\mathcal{E}|}\}$ be the set of all entities in the KB.
The TBox defines the schema structure of the KB including type ontology $\mathcal{T}$ and properties set $\mathcal{P}$.
The type ontology $\mathcal{T} = \{\tau_1,...,\tau_{|\mathcal{T}|}\}\subseteq \mathcal{E}$ consists of type definitions and a type hierarchy constructed with the \emph{subclass of} relation.
An entity $e \in \mathcal{E}$ can belong to one or more types of the type ontology. Specifically, we write $\mathit{type}(e)=\tau$ to denote an entity $e$ belongs to a type $\tau \in \mathcal{T}$.
The properties set $\mathcal{P} = \{p_1,...,p_{|\mathcal{P}|}\}$ defines the set of possible properties used to describe key attributes of an entity.
The ABox is the instantiation of KB which is composed of a set of RDF triples $\langle s, p, o \rangle$, where $s$ denotes a subject (an entity $e \in \mathcal{E}$), $p \in \mathcal{P}$ is a property (also known as predicate or relation) and $o$ denotes an object (an entity $e$, or a data value, e.g. number, time, string etc.).
We write $p(e)$ for \emph{property lookup} that returns $o$ when $\langle s,p,o \rangle$ exists in the KB, and $\bot$ otherwise. We implicitly assume that $p(e)$ maps to zero or one target.
Although tailored for our target KB, Wikidata\footnote{https://www.wikidata.org/}, the same notation applies to other knowledge bases like DBpedia and Freebase, etc.

\textbf{Definition 1.} Given table $T$ and entity set $\mathcal{E}$, \textit{table entity linking} aims to link entity mention $m_{k}$ in a specific cell $t_{i,j}$ of table $T$ to its referent entity in $\mathcal{E}$, or predict there is no corresponding entity in the KB, denoted $\bot$. In Figure~\ref{fig:pipeline}, \textcolor{blue}{1b} displays linked entities.

\textbf{Definition 2.} Given table $T$ and property set $\mathcal{P}$, \textit{property linking} associates a pair of columns, indexed by $s$ and $o$, with a property $p \in \mathcal{P}$ such that property $p$ relates $T_{[*,s]}$ and $T_{[*,o]}$ component-wise: $p(t_{i,s}) = t_{i,o}$ for all $i$. In Figure~\ref{fig:pipeline}, \textcolor{blue}{1c} displays linked properties.

\textbf{Definition 3.} Let $E = (e_{1, 1},  . . ., e_{n,1})$
be the list of entities corresponding to the left most table column, which we refer to as the \textit{main column}. \textit{Subject suggestion} generates a ranked list of entities $E_{\mathit{new}}$ to be added to $E$. In Figure~\ref{fig:pipeline}, \textcolor{blue}{2} displays suggested subjects.

\textbf{Definition 4.} Given table $T$ we extend the main column with the entity set $E_{\mathit{new}}$. For each $e_{i,1} \in E_{\mathit{new}}$ \textit{gap filling} returns $o_{i,j}$ where $o_{i,j}$ is either sourced from the KB via a triple $\langle e_{i,1}, p_j, o_{i,j} \rangle$, or via a verified web source. In Figure~\ref{fig:pipeline}, \textcolor{blue}{3} displays gap suggestions.

\textbf{Definition 5.} \textit{Row completion} is the task of returning a complete new row, that is, \textit{Row completion = Subject suggestion + Gap Filling}.

\section{Algorithms}
 
\begin{figure*}
  \includegraphics[width=1\textwidth]{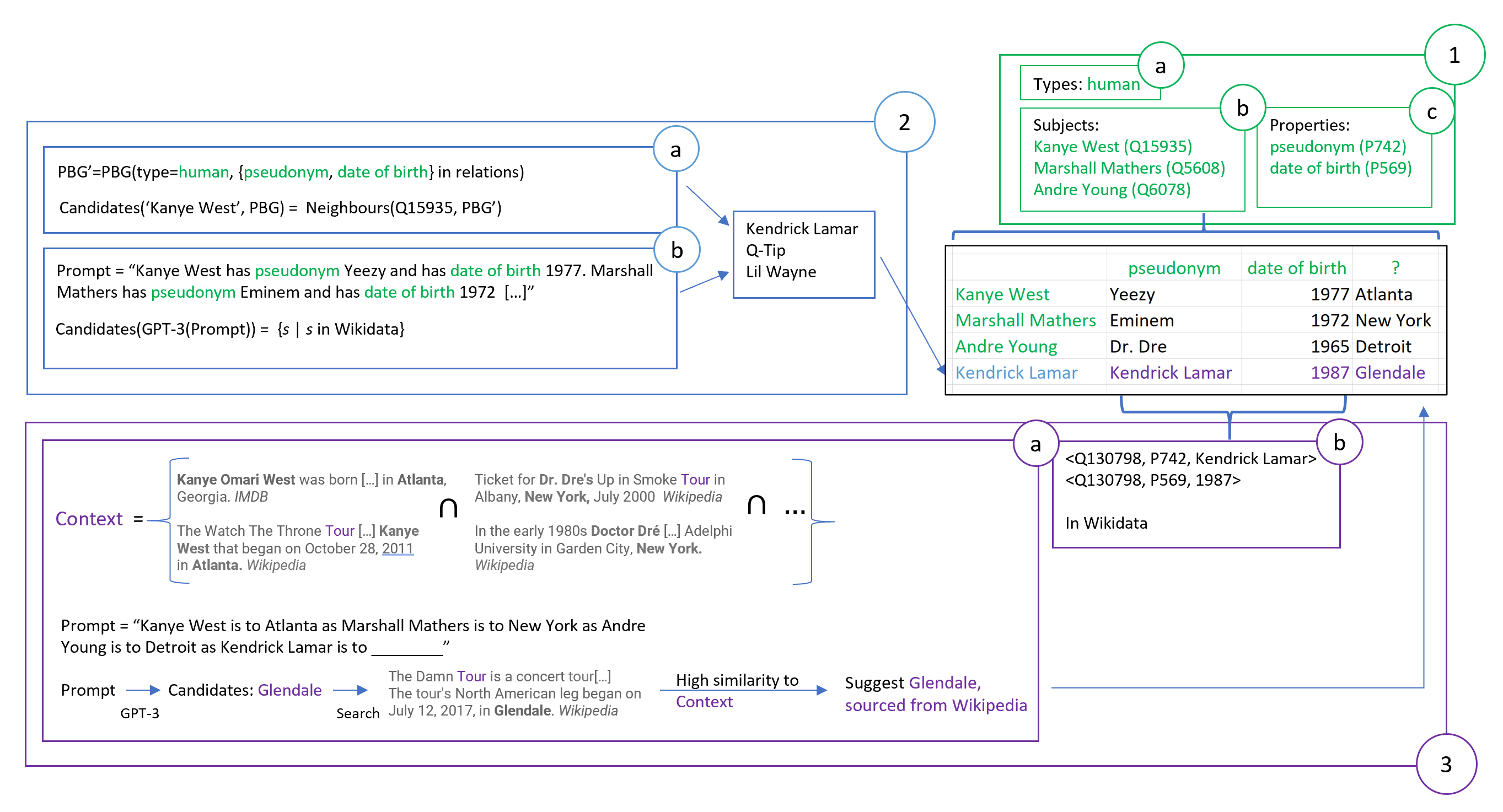}
  \caption{Row completion pipeline. In \textcolor{blue}{Step 1} we link the table to Wikidata, find the in-table properties (\textcolor{blue}{Step 1c}), and the possible types (\textcolor{blue}{Step 1a}) and corresponding Q-numbers for the subject entities (\textcolor{blue}{Step 1b}). To retrieve other subject entities (\textit{Kendrick Lamar}) we rank candidates generated via PBG' (\textcolor{blue}{Step 2a}) and GPT-3 (\textcolor{blue}{Step 2b}). To fill in the remaining relevant information we either suggest GPT-3 verified candidates (\textcolor{blue}{Step 3a}) or we retrieve directly from Wikidata (\textcolor{blue}{Step 3b}).}\label{fig:cg}
  \label{fig:pipeline}
\end{figure*}

Given a set of candidate entities for the main column we start by trying to find the most likely mapping between relational columns and Wikidata properties. We take a similar approach to \cite{bloomberg} with the exception of two problems that we address differently: property sparseness and numerical matching. Full details of this extension can be found in Appendix A and in this section we will focus on the algorithms for Subject Suggestion and Gap Filling. 

At the end of this phase our algorithm has produced the metadata shown in \textcolor{blue}{Step 1} in Figure~\ref{fig:pipeline}. We find the in-table properties \textit{pseudonym} and \textit{date of birth} and fail to link the third column. We manage to link all entities in the main column and identify possible types for them (such as \textit{human}). 

\subsection{Subject suggestion}
\begin{algorithm}
  \begin{algorithmic}
    \STATE $\mathit{suggest}(T, \mathcal{L}) =$ 
    \STATE \,\,\, $\mathcal{C} \leftarrow \{ e \mid \forall i < m, e\in\mathcal{E}. \,\, \mathit{distance}(\mathcal{L}_i,e) < \mathit{threshold} \}$ \textcolor{blue}{\circledNorm{2a}}
    \STATE \,\,\, $P \leftarrow \{ \mathit{link}(\mathit{GPT3}(\mathit{toPrompt}(T_{[i,*]}))) \mid \forall i < m \}$ \textcolor{blue}{\circledNorm{2b}}
    \STATE \,\,\, \textbf{return }$\mathit{rank}(\mathit{features}(T, \mathcal{L}, \mathcal{C} \cup P))$
  \end{algorithmic}
  \caption{Subject suggestion for table $T$ with seeds $\mathcal{L}$ (sketch)}
  \Description[subject suggestion algorithm]{formal representation of the algorithm}
  \label{algo:compl-rows}
\end{algorithm} 
Algorithm~\ref{algo:compl-rows} presents the high-level algorithm for subject suggestion which is represented by \textcolor{blue}{Step 2} in Figure~\ref{fig:pipeline}. 
Subject suggestion takes as input an $(m,n)$-table $T$ and the linked main column $\mathcal{L}$, and returns a ranked list of subjects to extend the main column.
The algorithm for subject suggestion comprises two sub-tasks: candidate generation and candidate ranking.

\paragraph{Candidate generation} Candidate generation is an important task as ranking all candidates in Wikidata has a high computational cost and at the current rate of growth it cannot scale even with the most efficient ranking algorithms. Our novel method to create a diverse candidate set is shown in Figure \ref{fig:pipeline}, \textcolor{blue}{Step 2}. 

First, we source KB candidates using the PBG embedding space (\textcolor{blue}{Step 2a}). Similar to \textit{Word2Vec}, PBG generates multi-relation embeddings by maximizing the (transformed) similarity score for existing RDF triples while minimizing it for non-existent ones. The similarity is measured by cosine distance (TransE) or dot product distance (RESCAL, DistMult, ComplEx) over transformed vectors for the source and object. The transformations considered are linear transformations (RESCAL), translations (TransE) and complex multiplication (DistMult, ComplEx). A viable candidate is a neighbor of at least one of the seeds present, shares at least one type in common with the seeds, and has some of the in-table properties. 

A key assumption in our work is that using nearest neighbors in PBG is meaningful for generation via similarity. We have found that requiring full property overlap is too restrictive for our choice of KB considering the property sparseness in Wikidata. In some cases (for example when the type is human) we need to further limit the number of candidate generations so we constrain by demanding all in-table properties to be present. If we have a small number of seeds (our current setup), we search for the nearest neighbors of each seed. We can extend this approach for very large tables when we might want to keep extra seeds by checking if we can find clusters in the seed space. Using PBG embeddings has a few important strong points. We did not have to train a new set of embeddings on relevant tabular data to build a space that is representative for our setup (as proposed in \cite{Li_Zhang}) because the embedding space is already constructed based on the relations between entities. We also do not have to worry about restricting our search to a fixed number of hops. By transitivity if the RDFs $ \langle s,p,o  \rangle $ and $\langle s,p,o'  \rangle$ exist then \textit{o} and \textit{o'} will likely be neighbors (and similarly for $\langle \mathit{s, p, o}  \rangle$ and $\langle \mathit{s', p, o}  \rangle$) as described in \cite{BigG}. As a final point, PBG updates efficiently with extra data, which is important as we want to have up-to-date suggestions. 

The second approach for candidate generation is to enhance the candidate pool with predictions obtained from GPT-3 free text generation (as shown in \textcolor{blue}{Step 2b} of Figure~\ref{fig:pipeline}). GPT-3 is an autoregressive language model with 175 billion parameters that has shown promising performance in the few-shot setting. Input table $T$ is converted to a prompt and used to sample GPT-3, from which candidates are generated. To create the prompt for GPT-3 we extractively summarize the seed rows given the properties identified. We create a template by concatenating the transformed $\langle \mathit{s, p, o}  \rangle$ $\rightarrow$ \textit{"s has p o"} that we identify on each row and we provide the text generator the template. The text generator then fills in new templates and we extract the entities from the templates. If we can link the entities to Wikidata we add them to the candidate set.

\paragraph{Candidate ranking} After we create a set of candidates from PBG ($\mathcal{C}$) and GPT-3 ($\mathcal{P}$) we want to create a ranked list. In practice, to maintain high levels of recall, the number of candidates we generate is considerably larger than the desired number of completions. 

In order to address the large class imbalance we rank by using the output of an outlier detector that returns the probability that a given candidate is part of the suggestion set. The assumption that we make is that \emph{given a set of well curated features the relevant suggestions will appear as outliers in the candidate set}. Our models make use of the following features:
\begin{enumerate}
 \item the distance to the closest seed in PBG (the choice of metric depends on the way the graph is generated e.g cosine similarity for TransE, dot product for ComplexE),
 \item the percentage of properties not in the table in common with seeds, 
 \item minimum normalized Levenshtein distance between candidate entity label and seed entity labels,
 \item the minimum string embedding (in FastText space \cite{FastT}) cosine distance between the candidate entity label and seed entity labels,
 \item the percentage of types shared between seeds and candidate,
 \item the percentage of seeds that have the candidate as a neighbor,
 \item the candidate generator (GPT-3, PBG or both),
 \item the GPT-3 generation score (if available). 
\end{enumerate}

We start from the toolbox proposed in \cite{pyod} for outlier detection algorithms. For semi-supervised and supervised systems we first use a kNN algorithm on the training data to assign table-clusters based on our feature space (and then map the test tables to a given cluster). As a supervised system we use XGBOD \cite{XGBOD}, an ensemble system that combines unsupervised outlier mining algorithms to extract representations to improve the feature space on top of which runs a supervised classifier. For each cluster we train a VAE \cite{VAE} on non-outliers (incorrect predictions) and find outliers using the reconstruction error (the hypothesis being that the higher the reconstruction error, the most likely a candidate is to be a viable suggestions). We also consider unsupervised methods that are proximity-based (e.g. LOF) or neural networks (e.g. MO-GAAL~\cite{MOGAAL}). 

\subsection{Gap filling}
\begin{algorithm}[h]
  \[
    \begin{array}{l}
      \mathsf{rankValues}(s, j, \mathcal{L}, p?) = \\
      \,\,\, \mathit{context} \leftarrow \mathsf{contextOfSeeds}(\{ \mathsf{bing}(\mathcal{L}_{[i,1]}, p, \mathcal{L}_{[i,j]}) \mid i \leq \mathsf{rows}(\mathcal{L}) \}) \\
      \,\,\, \mathit{scoredValues} \leftarrow \emptyset \\
      \,\,\, \text{\textbf{for} \textit{i} \textbf{in} \textit{sample}}\\
      \,\,\, \,\,\, \mathit{o} \leftarrow \mathsf{GPT3}(\mathsf{toPrompt}(\mathcal{L}_{[*,1]} \cdot \mathcal{L}_{[*,j]})) \\
      \,\,\, \,\,\, \mathit{snippets} \leftarrow \mathsf{bing}(s, p, o) \\
      \,\,\, \,\,\, \mathit{best} \leftarrow \mathsf{arg max}_{x \in \mathit{snippets}} \,\, \mathsf{score}(x, \mathit{context}) \\
      \,\,\, \,\,\, \mathbf{if} \, {\mathit{best_{\mathit{score}}} > \mathit{threshold}} \, \mathbf{then} \\
      \,\,\, \,\,\, \,\,\, \mathit{scoredValues} \leftarrow \mathit{scoredValues} \cup \{ (o, \mathit{best_{text}}) \} \\
      \,\,\, \mathbf{return} \,\, \mathit{scoredValues} \\[2ex]
      \mathsf{gapFill}(i, j, \mathcal{L}) = \\
      \,\,\, \mathit{s} \leftarrow \mathcal{L}_{[i, 1]}, \mathit{p} \leftarrow \mathsf{propertyLinks}(\mathcal{L}, j) \\
      \,\,\, \mathbf{if} \langle s, p, o \rangle \in \mathsf{KB} \, \text{\textbf{then return} $\{ (o, \langle s, p, o \rangle) \}$} \, \textcolor{purple}{\circledNorm{3b}} \\
      \,\,\, \mathbf{return} \, \mathsf{rankValues}(s, j, \mathcal{L}, p) \, \textcolor{purple}{\circledNorm{3a}}
    \end{array}
  \]
  \caption{Ranked Gap Filling for cell $(i,j$) in linked table $\mathcal{L}$.}
  \Description[formal description for the gap filling algorithm]{For every sample we first try to retrieve directly from the KB. If that is not possible we generate a GPT3 completion and then we look for snippets in Bing that contain the completion. We also look for a context that encompases the seed information and then compare the snippets and the context. If the similarity between the context and the snippets is high then we suggest the completion.}
  \label{algo:gap-filling-algo}
\end{algorithm}
Once we link the table to the KB and find good suggestions to append to the main column we can in principle easily fill in the remaining information in the table from the KB. Unfortunately, there are two cases where we cannot retrieve the information: when we cannot identify the property the column represents (for example, when the property does not exist in the KB, as shown in column four in Figure~\ref{fig:pipeline}) and when we can identify the property but the entity we are trying to fill the value for does not have the property in the KB (for example, an athlete's eye colour is not recorded).

In order to address these challenging situations we query GPT-3 for potential fills. To mitigate hallucinations we want to link the generation to at least one web source (as GPT-3 is trained on web data). Unfortunately, just linking is insufficient as we could find that although the fill is accurate it is not relevant for the table, for example GPT-3 could retrieve the correct hair colour of the athlete but we are interested in their eye colour. Our approach compares the context from the linked web source with a synthesized context that links the seed entities and their relevant values to decide if the fill is relevant.  Another benefit of linking is that it provides a way to mitigate the fact that GPT-3 might suggest out of date information (as the language model is unlikely to update frequently). For example, GPT-3 could retrieve the population of the US, but the data might not be from the most recent census. When linking to a web source we could warn the user the information is out of date from the metadata of the web source.

Algorithm~\ref{algo:gap-filling-algo} presents the algorithm for ranked gap filling which is represented by \textcolor{purple}{Step 3} in Figure~\ref{fig:pipeline}.

Formally, gap filling takes as input a target cell $(i,j)$ and a \emph{linked} $(m,n)$-table $\mathcal{L}$, and returns a set of ranked values with provenance. Provenance is either an RDF-triple or a sourced text snippet.

Gap filling proceeds by obtaining the subject $s$ for the row containing the gap, and the linked property $p$ for the column containing the gap. Property $p$ can be undefined when property linking fails, for example because the underlying KB lacks the relation.
When the KB contains a triple with the corresponding subject and property we immediately return this triple as the value to populate the gap, as shown in \textcolor{purple}{Step 3b} in Figure~\ref{fig:pipeline}.
If a triple does not exist (because the property or value is missing) we query GPT-3 to retrieve a set of candidates, corresponding to function $\mathsf{rankValues}$.

The function $\mathsf{rankValues}$ takes as input a subject $s$, a
column $j$, a \emph{linked} $(m,n)$-table $\mathcal{L}$, and an
optional property $p$ denoted by $p?$.
First we define $\mathit{context}$, a loose context given the seed information. For each seed row we query Bing for representative sentences using the seed subject, the value in column $j$ of the row, and optional property $p$ as keywords by using the $\mathit{bing}$ function. We then use sentence transformers to encode the information into a context. To find the most likely context shared among the seeds we compute the (cosine) similarity across encodings. All the sentences that are most similar (with similarity above a learned threshold) become the context. This process is defined by $\mathit{contextOfSeeds}$. 

The next step is to iteratively sample GPT-3 for gap suggestions using a prompt created by concatenating prompts for every seed row. When the property $p$ is missing we use an analogy-based prompt to retrieve a set of candidates, as shown in \textcolor{purple}{Step 3a} in Figure~\ref{fig:pipeline}. For a subject $s$ and value $o$ we construct prompt ``\emph{s} is to \emph{o} as''.
When the property $p$ is linked we construct a prompt of the form ``\emph{s} has \emph{p} \emph{o}''. The function \emph{GPT3} returns the GPT-3 suggestions that are obtain by using the prompt generated by \emph{toPrompt}.
We prune incorrect GPT-3 candidates by verifying against information from trustworthy sources, producing \emph{snippets}, a set of sourced text snippets. Sourced snippets are obtained by linking candidate $\langle s, p, o \rangle$ triple to Wikipedia or news articles using a web-search engine (like the Bing API).
Each source snippet is scored using (cosine) similarity to \emph{context}, and if the best snippet has sufficient score, the value and source is added to \textit{scoredValues}.

Unlike prior approaches, our solution does not require having a two step process to first determine if a cell should be filled. We learn a rough threshold on the validation set such that candidates with scores less than the given threshold are not proposed which can lead to blank cells. 

\section{Experimental Evaluation}
\label{sec:exper-eval}
In this section we present our experimental results. By convention, in every table of results we indicate prior work using a citation, and all other entries are results of this work.

\subsection{Details for reproducibility}

\paragraph{Datasets}

In this work we run extensive experiments on the standard benchmark proposed in \cite{entitables}, a dataset of 1000 real tables curated by humans. Most importantly this dataset is suitable for evaluation for candidates sourced either from DBpedia or Wikidata as the tables are extracted from Wikipedia (so either source should be able to retrieve all relevant suggestions). To our knowledge there are currently no benchmarks that are more extensive. 

As the WikiTables dataset was curated in 2014 as part of TabEL \cite{tabel}, the set of ground truth suggestions can be incomplete (for example, the table about US presidents does not include Donald Trump), or some of the values are out of date (for example, the data is from a previous census). Unfortunately, extending the ground truth tables can be fairly noisy so we chose a more conservative approach: we created a set of rough table extensions by scraping Wikipedia and we checked if any of the extra entities are among the highly ranked candidates in our model. After manual validation we removed them when reporting results. We have chosen this approach in order to fairly compare with prior art which uses the data from the 2016 version of DBpedia. This problem affects at least 67 out of 1000 tables.

A second issue with the curated test set is that the main assumption in the task definition does not necessarily hold for this dataset. A significant number of tables in the test set are not strictly relational (as defined in \cite{survey}). We estimate that at least 83 tables in our dataset are not strictly relational, but a join of several relational tables. To align with prior art and the standard definition of subject suggestion, we report results relative to the first column even though more properties relate to another column. 

For rapid development we created a pipeline to download Wikidata dumps and load them into a Spark distributed cluster in DataBricks that accepts between 2-11 workers (Standard DS4 v2 with 28GB Memory, 8 Cores, 1.5 DBU). 
The full pipeline can efficiently run in Databricks and the runtime upper-bound for the full test set is 14 minutes for entity-property linking, 52 minutes for candidate generation, 11 minutes for feature generation, 6 minutes for ranking and 58 minutes for gap filling. We believe the runtime can be significantly improved as we have not focused on optimizing the pipeline. For reproducibility we use the pre-trained PBG embeddings for Wikidata\footnote{https://github.com/facebookresearch/PyTorch-BigGraph} (which includes 78 million entities), the 300 dimensional, 1 million FastText Embeddings\footnote{https://fasttext.cc/docs/en/english-vectors.html} trained with subword information on Wikipedia 2017, UMBC web base corpus and statmt.org news dataset and the Hugging Face sentence transformer with bert-base-nli-mean-tokens\footnote{https://huggingface.co/sentence-transformers/bert-base-nli-mean-tokens}. 

\paragraph{Particulars to reproduce Candidate Generation Experiments}
In order to generate GPT-3 candidates we call the API\footnote{https://beta.openai.com/} 100 times (with 1 sentence completions) and use temperature 0.7 (we trade stability for diversity). We generate 100 raw candidates per table out of which, on average, 73 can be linked to a Wikidata entity.  

For PBG generation imposing type constraints reduces the search space by a factor of 50 at the cost of limiting maximum average recall to 93.8\%. Introducing the in-table property constraint reduces the space further by a factor of 2 and improves recall slightly.

\paragraph{Particulars to reproduce Ranking Experiments}

In our work we have experimented with the available detectors in \cite{pyod}. In general we found that ensembles give a performance boost of 5\%-7\% and in particular XGBOD \cite{XGBOD} and LSCP \cite{LSCP} perform best. We set the contamination level between 1\%-6\% and as we use the output just for ranking (and not classifying), we are looking for methods that are tolerant to changes in the contamination hyperparameter. For proximity based methods we are also interested in the number of neighbours and for all the methods we have tried, the results did not change drastically by changing this hyperparameter. 

For unsupervised methods, MO-GAAL\cite{MOGAAL} outperformed other neural network approaches, but we found that it was not particularly stable for our set of features (level of contamination and number of subgenerators had a fair impact).

\paragraph{Particulars to reproduce Gap Filling Experiments}

For row completion we first try to retrieve the relevant RDF triples. If that is not possible we query GPT-3 under the setup previously described for candidate generation. Using the output of the gap filling generation we proceed to try to verify by linking to web sources. We call the Bing API\footnote{https://www.microsoft.com/en-us/bing/apis/bing-web-search-api} restricted to news and Wikipedia articles and we retain the top 10 matches. We construct the description of the page by concatenating the \textit{name} and \textit{snippet} fields. In our experiments we found that using the ranking of the search engine was not fruitful. To extend the search one can use the \textit{deep links} option to return related webpages that Bing found on the webpage’s website or the \textit{relatedSearch} option to return a list of most related queries made by other users. For our dataset this extension did not improve results. We threshold our results by showing a fill only if the source similarity score is above 0.05.

\subsection{Experimental results}
\paragraph{Candidate generation results}

Unlike previous approaches we generate candidates from Wikidata and unfortunately the prior generation algorithms are not suitable for this KB. Categories play an important role in previous approaches \cite{entitables}, but for the vast majority of Wikidata entities categories are not provided and thus we cannot fairly compare approaches by a direct implementation of prior art. Recent work extends categories to Wikidata \cite{categ} and in our future work we will investigate leveraging this approach.

We provide a first comparison in Table \ref{table:cand_nr_gen} by looking at the performance of our approach versus the current state of the art in candidate generation, EntiTables\cite{entitables}. For KB generation we consistently improve over prior art despite searching in a significantly larger space. When introducing other sources (tables versus free text) we perform significantly better for small numbers of candidates, but our advantage reduces as we increase the number of generations. We hypothesize that the discrepancy in KB size is relevant when generating many candidates. In practice, ranking significantly more than 1000 candidates is not practical so our goal is to have quality generations within 1000 candidates.  
\begin{table}[t]
  \begin{center}
    \begin{tabular}{ |l|l|l|l|l| }    \hline
      \multicolumn{4}{|c|}{Average Recall per nr candidates generated}             \\ \hline
      \textsc{Method} & \textsc{Anchor} & \multicolumn{2}{|c|}{\# Candidates}\\ \hline
      \textsc{} & \textsc{} & \textsc{1000 } & \textsc{10000} \\ \hline
      EntiTables KB~\cite{entitables} & DBpedia      & 64.1 & 78.8 \\ \hline
      Entitables~\cite{entitables} & + tables & 78.4 & 94.0        \\ \hline
      PBG TransE/ComplexE     & Wikidata & 72.2/72.4  &  88.1/88.2  \\ \hline
      + GPT-3 (Our method)   & + free text  & 87.4/87.5 & 94.5/94.7 \\ \hline
    \end{tabular}
    \vspace{0.1cm}
    \caption{Candidate generation for subject suggestion. Average recall for generations from top 3 seeds. For 1000 generations we observe a significant gain over prior art. The difference between PBG versions is not significant.}
    \Description[Candidate generation for subject suggestion]{The table shows the average recall for generations from top 3 seeds. For 1000 generations we observe a significant gain over prior art. The difference between PBG versions is not significant.}
    \label{table:cand_nr_gen}
  \end{center}
  \vspace{-0.5cm}
\end{table}

Previous approaches only report results by generating suggestions when they consider the top-k rows as seeds. We argue that unless we treat rows as an ordered collection the generation should be order agnostic. We report average recall for 3 or 4 seeds in Table \ref{table:cand_gen} and generate 5000 candidates. We have included only 3 and 4 seeds as previous models have peak performance in this case and GPT-3's prompt requires a fair amount of information. The error margin is relatively small for PBG and prior art, but significant for GPT-3. In our study we observed that GPT-3 performs very well for tables when there is an apparent order (for example, consecutive years). Our best hypothesis is that when the seeds have a certain pattern GPT-3 confidently reproduces the pattern, which is consistent with previous studies \cite{GPT3}. By improving the prompt we have gained a significant performance boost while shrinking the variation. When we tried the same approach with GPT-2\footnote{https://huggingface.co/gpt2} we did not achieve satisfactory results. We have ran secondary experiments on an annotated 100 instances, where we noticed that GPT-3 is resilient to badly worded prompts (i.e. that are not fluent), but GPT-2 predictions improved (by 8\%). Based on our experimentation around various prompt modifications we conclude that this approach does not generalize well for less performant language models. We also explored using Codex\footnote{https://openai.com/blog/openai-codex/}, a descendant of GPT-3 for code generation. We encoded the table as a Pandas DataFrame and included it in Codex's prompt together with the ask to add more examples. The relevance of the generations decreased, but the structure quality improved (i.e. more suggestions had the correct type). We hypothesize that as Codex is trained on a significant amount of data from Github, and most examples contain toy datasets, it loses some of its capability to \textit{directly} retrieve factual information. 
\begin{table}[t]
  \begin{center}
    \begin{tabular}{ |l|l|l| }    \hline
      \multicolumn{3}{|c|}{Stability study}             \\ \hline
                                  & \textsc{3 seeds}      & \textsc{4 seeds}      \\ \hline
      Entitable KB~\cite{entitables} & 73.1 +/- 0.1 & 74.9 +/- 0.08 \\ \hline
      Entitables~\cite{entitables}  & 91.6 +/- 0.08 & 92.5 +/- 0.07 \\ \hline
      PBG               & 76.2 +/- 0.09 & 77.6 +/- 0.08       \\ \hline
      GPT-3 base       & 43.8 +/- 2.1 & 46.5 +/- 3.2        \\ \hline
      GPT-3 best & 59.7 +/- 1.5 & 61.4 +/- 1.7 \\ \hline
      GPT-2 & 31.1 +/- 3.6 & 32.9 +/- 3.3 \\ \hline
      Codex & 49.2 +/- 2.0 & 48.4 +/- 1.9 \\ \hline 
      PBG + GPT-3 best  (Our method)   & \textbf{94.1 +/- 0.5} & \textbf{94.3 +/- 0.4}        \\ \hline
    \end{tabular}
    \vspace{0.1cm}
    \caption{Subject suggestion. Average recall performance with error-bars by choosing every 3-seed/4-seed combination in top 5 rows. We limit to an average of 5000/6000 candidates per table to recover the EntiTables generations from \cite{Li_Zhang}. GPT-3 base uses as prompt the table as is, GPT-3 best uses our improved prompt as shown in Figure 1, GPT-2 uses the same prompt as GPT-3 best and Codex uses as prompt the table as a Pandas DataFrame alongside a simple ask.}
    \Description[Subject suggestion stability study]{The table shows average recall performance with error-bars by choosing every 3-seed/4-seed combination in top 5 rows. We limit to an average of 5000/6000 candidates per table to recover the EntiTables generations. GPT-3 base and Codex which use as prompt the table have very high variability within +/-2, but when we combine them with PBG we get close to the low variability from Entitables. We have 0.5 they have 0.1.}
    \label{table:cand_gen}
  \end{center}
  \vspace{-0.5cm}
\end{table}

\paragraph{Ranking results}

For the task of table completion we test the capabilities of the ranking function by using the candidates previously generated. When comparing the generation results between the two methods we find that in the case where we restrict candidate numbers to 5000 we have only 1122 candidates in common (on average). As we want to test the ranking function we run our outlier detection algorithms over both candidate sets in Table \ref{table:row_comp}. As baselines we include Table2Vec~\cite{Li_Zhang} which augments the approach from EntiTables~\cite{entitables} with static table embeddings, as well as a new enhancement of EntiTables with a new, deep learning tabular representation TABBIE\footnote{using the implementation found at https://github.com/SFIG611/tabbie}~\cite{tabbie}. 

 In our work we report results only for 3 or 4 seeds as we found that the results for 2 and 5 seeds are very similar (in line with the findings of prior art). The first two rows in the table show that by extending the baseline with TABBIE's table representation we can improve upon Table2Vec. On its own, TABBIE ranks 300k entities and has a performance of 43.4/44.1 for 3/4 row seeds. When coupling it to our approach or EntiTables we restrict the classifier to the set of candidates we are considering. 

\begin{table}[t]
  \begin{center}
    \begin{tabular}{ |l|l|l|l|l| }    \hline
      \multicolumn{5}{|c|}{Mean average precision for subject suggestion}             \\ \hline
        
      \textsc{Method}  &  \multicolumn{2}{c|}{\textsc{Our candidates}}
            & \multicolumn{2}{c|}{\textsc{EntiTables}}\\ \hline
      &\textsc{3}& \textsc{4}&\textsc{3}&\textsc{4} \\ \hline
      Table2Vec~\cite{Li_Zhang} & 55.8 & 56.7 & 64.0 & 65.2 \\ \hline
      EntiTables~\cite{entitables} + TABBIE~\cite{tabbie} & 63.2 & 63.7 & \textbf{66.0} & \textbf{66.2} \\ \hline
      XGBOD (Our method)  & 72.4 & 72.6 & 59.9 & 60.2   \\ \hline
      XGBOD + TABBIE ~\cite{tabbie} & \textbf{72.8} & \textbf{72.9}& 64.9 & 64.9 \\ \hline
    \end{tabular}
    \vspace{0.1cm}
    \caption{Ranking for subject suggestion. Mean average precision (MAP) for subject suggestion by taking the top 3/4 rows as seeds. We report performance on the whole ground truth set and inherit a handicap from candidate generation (by comparing performance from our candidates and EntiTable's).}
    \label{table:row_comp}
  \Description[Ranking for subject suggestion]{We compute mean average precision (MAP) for subject suggestion by taking the top 3/4 rows as seeds. We report performance on the whole ground truth set and inherit a handicap from candidate generation (by comparing performance from our candidates and EntiTable's). Our best result is 72.8 compared to Entitable's best result of 66}
  \end{center}
  \vspace{-0.8cm}
\end{table}
In our work we have experimented with several outlier detectors including VAEs\cite{VAE}, MO-GAAL\cite{MOGAAL} and LSCP+LOF\cite{LSCP}, but found they perform 5-10\% worse than XGBOD\cite{XGBOD}. One common result across all detectors is that they significantly over perform for our candidate set compared to the one from EntiTables. We hypothesize this is caused by the way we constructed our feature space to take advantage of PBG's/GPT-3 structure which does not translate as well for DBpedia entities mapped in the Wikidata space. 

As shown in Table \ref{table:row_comp}, we achieve best results when we include TABBIE's classifier scores for our candidates as an extra feature for XGBOD. We observe that the performance increase is not significant as the added feature is highly correlated with GPT-3's generation score (a Pearson Coefficient of 0.89) and medium correlated with the distance to closest seed in PBG (a Pearson Coefficient of 0.72). 

Thus, we conclude that for subject suggestion using a language model to generate candidates has a significantly bigger impact than using it to encode a table when combined with KB mining. On tables that are not relational this conclusion is unlikely to hold and we leave it to future research to investigate other types of tables. The choice of language model can be impactful and our experiments suggest that the capacity of the model, and the data the model is trained on are important.
 
\paragraph{Gap Filling}

For gap filling we compare against various completion methods as shown in Table \ref{table:row_c}. TURL \cite{deng2020turl} improves gap filling with tabular data over prior art CellAutocomplete \cite{cac} and we use it as a baseline. Unlike CellAutocomplete we first try to link to a KB and if that is not possible we use auxiliary methods. By including GPT-3 suggestions our method significantly improves over the direct KB retrieval baseline. Unsurprisingly, as the tables are sourced from Wikipedia we find that 82\% of GPT-3 suggestions are grounded in Wikipedia but we also find that 37\% can come from other sources as well (for example, BBC). As the dataset is fairly simple we find very few cases (5\%) when GPT-3 generates more than 2 viable completions and it mainly generates a single viable completion (78\%). An interesting case where the GPT-3 completion fails to return reasonable candidates is for numerical cells. The recall for numerical cells is 38.4\% in contrast with strings for which the recall is 82.9\%. By looking at the suggestions the language model returns numerical values that are of the right type (for example, it returns years when appropriate) but are hallucinations. In this dataset most numerical cells were linked by the KB, but we think that including table candidates would be beneficial in other datasets.  

To link a GPT-3 completion to an article we must firt generate a loose context from the seed rows. To do so we have experimented with bag of words overlap and standard sentence embeddings as well as sentence transformers. The first two methods had similar performance and led to 78\% precision and we obtained a slight improvement by introducing transformers (79.3\%). 

\begin{table}[t]
  \begin{center}
    \begin{tabular}{ |l|l|l|l|l|l| }    \hline
      \multicolumn{5}{|c|}{Precision/Recall Gap Filling}             \\ \hline
        
      \textsc{Method} & \textsc{Anchor} & \textsc{P @ 1} & \textsc{R @ 1} 
      & \textsc{P @ 3}  \\ \hline
      KB  & Wikidata (Wd) & \textbf{97.8} & 37.2 & \textbf{97.8}    \\ \hline
      TURL~\cite{deng2020turl} & tables & 49.2 & 41.1 & 66.3  \\ \hline
      TURL + KB~\cite{deng2020turl} & Wd + tables & 63.8 & 55.3 & 76.1 \\ \hline
      GPT-3 + KB (Our) & Wd + free text & 79.3 & \textbf{70.2} & 82.2  \\ \hline
    \end{tabular}
    \vspace{0.1cm}
    \caption{Gap Filling. We extend prior methods by including candidates that are verified in news or Wikipedia articles and improvements in both recall and precision are significant (around 15\%). The test set includes all cells, except the subject column, for non-seed rows in the 1000 WikiTables dataset.}
    \Description[Gap Filling]{In this table we extend prior methods by including candidates that are verified in news or Wikipedia articles and improvements in both recall and precision are significant (15.5 for precision, 14.9 for recall). The test set includes all cells, except the subject column, for non-seed rows in the 1000 WikiTables dataset.}
    \label{table:row_c}
  \end{center}
  \vspace{-0.8cm}
\end{table}

\section{Conclusions and Future Work}

The main goal of our work was to provide a system that successfully suggests new, complete rows from large, diverse sources like Wikidata or trustworthy websources. 

We showed that despite metadata scarcity (i.e no headers or captions) we can start from a few examples and generate the relevant metadata to a quality that is good enough to exploit by pre-trained systems such as GPT-3 or PBG. 

From our studies we conclude that candidate generation is an extremely important step and language models can play a key role in improving this space. For relational tables the gains from having a rich tabular representation (that comes at a significant cost) can be surpassed by generating a high-quality set of candidates.

\paragraph{Limitations} Although our work has shown great potential there are still limitations that should be addressed. The current scope, relational single subject tables, is quite restrictive and further research needs to be conducted to first extend to multi subject tables and then to mixed tables (not strictly relational). As discussed in the results section we have noticed performance drops significantly for numerical properties (including dates) and we need to improve the system for such cases. Finally the system is tailored for Wikidata, but it should be extended to other KBs as well. 

\paragraph{Future versions} An interesting direction for future work is to create a system that holistically improves subject suggestion and row completion simultaneously, for example by creating a feedback loop. We plan to investigate this option in our system's second version. Before releasing such a system in the wild, we would like to explore how we can extend our approach to be user-centric and define new measures for success.

Unfortunately, without better datasets it is hard to validate that new approaches actually improve upon prior art. Improving current datasets is a priority for our work and we will investigate annotating a large corpora.

\clearpage

\ifappendix
\appendix

\section{Acknowledgement}
We would like to thank Christian Canton, Chris Oslund, Nick Wilson, Lena Yeoh and Yordan Zaykov for the insightful conversations, design jams and overall support throughout our project.

\section{Property linking}
\label{PropertyLinking}

Given a set of candidate entities for the main column we try to find the most likely mapping between relational columns and Wikidata properties. We take a similar approach to \cite{bloomberg} with the exception of two problems that we address differently: property sparseness and numerical matching.

In this section we first describe our generic property linking
function (Algorithm~\ref{algo:compl-prop-link}), and then we describe
our type-specific scoring functions for numeric and string values.

\begin{algorithm}[h]
  \[
    \begin{array}{l}
      \mathsf{link}(T, \mathcal{L}, j) = \\
      \,\,\, \mathcal{P} \leftarrow \{p \in \mathcal{P} | \exists e \in \mathcal{L}. \, p(e) \not= \bot \} \\
      \,\,\, \mathrm{scores} \leftarrow \{  p \mapsto \sum_{i \leq m} \mathrm{score}_{ij}(p) \mid p \in \mathcal{P} \} \\
      \,\,\, \mathrm{scores}_{\mathrm{max}} \leftarrow  \{ p \mapsto n \in \mathrm{scores} \mid n = \mathrm{max}(\mathrm{image(\mathrm{scores}))} \} \\
      \,\,\, \mathrm{scores}^{\approx} \leftarrow \{ p \mapsto n + n' \mid \forall p \mapsto n \in \mathrm{scores}_{\mathrm{max}}. \\
      \quad \, n' = \sum_{i \leq m} \text{if } \mathrm{score}_{ij}(p) = 0 \text{ then } \mathrm{score^*}_{ij}(p) \text{ else } 0 \} \\
      \,\,\, \mathrm{scores}^\approx_{\mathrm{max}} \leftarrow \{ p \mapsto n \in \mathrm{scores}^\approx \mid n = \mathrm{max}(\mathrm{image(\mathrm{scores}^\approx))} \}
    \end{array}
  \]
  \vspace{-2ex}
  \begin{algorithmic}
    \STATE \,\,\, \textbf{if} $\mathrm{scores}_{\mathrm{max}} = \{ p \mapsto n\} \land n \geq \mathit{threshold}$ \textbf{then return} $p$
    \STATE \,\,\, \textbf{if} {$\mathrm{scores}^\approx_{\mathrm{max}} = \{ p \mapsto n\} \land n \geq \mathit{threshold}$} \textbf{then return} $p$
    \STATE \,\,\, \textbf{else} $\mathsf{fail} \text{ ``Cannot Resolve Confidently''}$
  \end{algorithmic}
  \caption{Property linking for column index $j$}
  \label{algo:compl-prop-link}
\end{algorithm}

\paragraph{Generic Property Linking}
Algorithm~\ref{algo:compl-prop-link} presents our generic property
linking function $\mathsf{link}(T, \mathcal{L}, j)$. The function
$\mathsf{link}(T, \mathcal{L}, j)$ accepts as input an ($m, n$)-table
$T$, linked main column $\mathcal{L}$, and target column index $j$;
the function returns the linked property for column $j$ or fails.  We
write $x \leftarrow e$ to denote the assignment of the value of
expression $e$ to identifier $x$.

The property linking function is
implicitly parametrised by two scoring functions that assign a score
to a property $p$ for a particular cell $(i,j)$: we write
$\mathrm{score}_{ij}(p)$ for the \emph{exact} property score, and we
write $\mathrm{score^*}_{ij}(p)$ for the \emph{approximate}
property score. The implementation of $\mathrm{score}$ and
$\mathrm{score^*}$ depends on the type of the column we are linking:
numeric or string. The concrete implementations of these functions is
described later in the section.

Function $\mathsf{link}(T, \mathcal{L}, j)$ proceeds as follows.
First, define $\mathcal{P}$ to be the set of properties such that
there exists an entity in main column $\mathcal{L}$ with that
property.
Define $\mathsf{scores}$ to be the set of property-score
($p \mapsto n$) mappings where $p$ is in $\mathcal{P}$, and $n$ is the
sum of exact property scores for all values in the target column.
Define $\mathsf{scores}_{\mathrm{max}}$ to be the set of
property-score ($p \mapsto n$) mappings where $n$ is the largest score
in $\mathsf{scores}$. Multiple properties may share the highest score,
hence we record a set. Define
$\mathrm{image}(M) = \{ n \mid (p \mapsto n) \in M\}$.
Values as written in the table will frequently differ from the exact value as
defined in the KB, hence we also compute an approximate set of scores.
Define $\mathsf{scores}^\approx$ to be the set of property-score
($p \mapsto n + n'$) mappings where $p$ is in $\mathcal{P}$, $n$ is
the exact score, and $n'$ is the approximate adjustment. The
approximate adjustment is defined as the sum of approximate property
scores for all values in the target column that did not match exactly.
Define $\mathsf{scores}_{\mathrm{max}}^\approx$ analogously to
$\mathsf{scores}_{\mathrm{max}}$.
Finally, we determine the property link using
$\mathsf{scores}_{\mathrm{max}}$ and
$\mathsf{scores}_{\mathrm{max}}^\approx$.  When
$\mathsf{scores}_{\mathrm{max}}$ has a unique high score (is a
singleton set $\{ p \mapsto n\}$), and provides sufficient coverage
(determined by \emph{threshold}), we link using $p$. When that fails,
we apply the same process to $\mathsf{scores}_{\mathrm{max}}^\approx$,
and if approximate matching fails, we return no match.

\paragraph{Numeric Properties}
Matching numerical-valued columns to properties is an interesting problem as standard approaches for string matching (like fuzzy matching) are not as effective. For example, consider a column about the heights of athletes. As these values can vary between measurements and various sources could report different values, linking directly to a KB would not be possible. 

To address such challenges if direct matching methods are not fruitful (i.e. we cannot directly match within unit conversions) we consider the values in the columns holistically and we compare their statistics (for this paper we only consider ranges) with the statistics of the numerical properties for a given class of entities. For instance if we identify that the main column has type "athlete" we check if the values in the column match any of the numerical property values in the KB that are representative for athletes. If that is not the case we compare the statistics of the values in the column against reasonable statistics for numerical properties of athletes (e.g. if the range of our data is within 102-210 it is consistent with heights for athletes).

 Formally, for numeric properties we introduce the concept of \emph{characteristic ranges}. Write $\mathcal{C}_{p;\tau}$ for the characteristic range of property $p$ for type $\tau$, defined as $\mathcal{C}_{p;\tau} = \mathrm{range}(\mathrm{characteristic}(\mathcal{E}_{p;\tau}))$, where $\mathcal{E}_{p;\tau} = \{p(e) | \forall e \in \mathcal{E}. \, \mathit{type}(e)=\tau, \, p(e) \not=\bot \}$ and \textit{p} is a numeric property.

The \textit{characteristic} function removes the outliers from the set by using the Isolation Forest Algorithm \cite{isofor}. The \textit{range} function returns upper and lower bounds for a set of numbers.

We now define a property scoring function for a numeric cell value $t_{i,j}$ in table $T$, given linked main column $\mathcal{L}$ which is a column vector of entities. First, write $\mathbbm{1}$ for the \emph{indicator} function that maps true to 1 and false to 0.
Define
$\mathrm{score}_{ij}(p) = \mathbbm{1}(\mathit{conv}(p(\mathcal{L}_i) =
t_{i,j}))$. We say that a cell value ($t_{ij}$) is consistent with the
property value of the linked entity in the same row ($\mathcal{L}_i$)
if their values are equal within unit conversion (\textit{conv}).
Define $\mathrm{score^*}_{ij}(p) = \mathbbm{1}(\exists \tau. \, \mathit{conv}(t_{i,j}) \in \mathcal{C}_{p;\tau} \wedge \mathit{type}(\mathcal{L}_i) = \tau)$. We say that a cell value ($t_{ij}$) is approximately consistent with the
property value of the linked entity in the same row ($\mathcal{L}_i$)
if their values are within the characteristic range, modulo unit conversion (\textit{conv}).

\paragraph{String Properties}

Wikidata has significant property sparseness since users are not obliged to include all properties when adding entities. This makes assigning properties to columns more challenging as direct or fuzzy string-matching can be insufficient. For example, if a column is about the eye colour of athletes we are likely to struggle to identify the property as a significant amount of athlete entities do not have eye colour as a property. 

 We want to address the case where we find that at least one value in the column matches to a property value, but other entities do not have that property and thus we cannot confidently assign the column. Our method estimates how likely it is that the property is missing (but should be present) for the other entities by checking if similar entities have it. In our previous example if most athletes in our table do not have the property eye colour, but similar athletes to them do, we can increase our confidence that eye colour is the correct match for our column. 

We define a property scoring function for a string cell value $t_{i,j}$ in table $T$, given linked main column $\mathcal{L}$.
Define $\mathrm{score}_{ij}(p) = \mathbbm{1}(p(\mathcal{L}_i) \approx t_{i,j})$. We say that a cell value ($t_{ij}$) is consistent with the property value of the linked entity in the same row ($\mathcal{L}_i$) if their values are equal within fuzzy matching, written $\approx$ (similarly to \cite{bloomberg}).
Define $\mathrm{score^*}_{ij}(p) = \operatorname{arg\,max}_{p \in pred} \max_{e \in \mathcal{E}_{1i}}\mathrm{score}(p,e)$.
We estimate how likely it is that the property is missing. We do this by looking at the properties of the $n$ nearest neighbors in PBG that share the type of $e$.

Similar to \textit{Word2Vec}, PBG generates multi-relation embeddings by maximizing the (transformed) similarity score for existing RDF triples while minimizing it for non-existent ones. The similarity is measured by cosine distance (TransE) or dot product distance (RESCAL, DistMult, ComplEx) over transformed vectors for the source and object. The transformations considered are linear transformations (RESCAL), translations (TransE) and complex multiplication (DistMult, ComplEx). 

Thus, for TransE embeddings we compute a score to estimate what is the likelihood that entity \emph{e} is missing property \emph{p} as $\mathrm{score}(p,e) = \overline{\sum_n} sim(e_i, e)(\mathbbm{1}(p(e_i) \not= \bot) -\mathbbm{1}(p(e_i) = \bot))$, where $sim(e, e_i) = 1-\frac{L_2(e, e_i)}{\max(\{L_2(e, e_j)|j<n)\})}$ and where the operator $\overline{f}$ acting on function $f$ is defined as $\overline{f}(x) = \min(\max(0, f(x)), 1)$.

\paragraph{Entity-Property linking results}
In our work we extend a previous system \textit{LinkingPark} that was recently externally evaluated in the SemTab Competition. The system ranked 2nd overall performing particularly well for the main entity type annotation task (CTA)\footnote{https://www.cs.ox.ac.uk/isg/challenges/sem-tab/2020/results.html}. 
By enhancing the prior pipeline with the new property linking module we improve performance slightly over the SemTab dataset (unfortunately for this task simple methods like direct linking achieve around 95\% precision). For Rounds 1-3 we have an up to 1\% improvement over the base model and for Round 4 a 1.2\% improvement. The main performance difference can be seen for the WikiTables dataset. By including the new module we improve coverage from 38\% to 44\%. This dataset does not have ground truth available for KB linking but we can infer from the gap filling performance via RDFs that precision is 98.7\%.
\fi

\bibliographystyle{ACM-Reference-Format}
\bibliography{biblio}

\end{document}